# Sentence, Phrase, and Triple Annotations to Build a Knowledge Graph of Natural Language Processing Contributions —A Trial Dataset

Jennifer D'Souza[†], Sören Auer

TIB Leibniz Information Centre for Science and Technology, Hannover, Germany

**Abstract**

**Purpose:** This work aims to normalize the NLPCONTRIBUTIONS scheme (henceforward, NLPCONTRIBUTIONGRAPH) to structure, directly from article sentences, the contributions information in Natural Language Processing (NLP) scholarly articles via a two-stage annotation methodology: 1) pilot stage—to define the scheme (described in prior work); and 2) adjudication stage—to normalize the graphing model (the focus of this paper).

**Design/methodology/approach:** We re-annotate, a second time, the contributions-pertinent information across 50 prior-annotated NLP scholarly articles in terms of a data pipeline comprising: contribution-centered sentences, phrases, and triple statements. To this end, specifically, care was taken in the adjudication annotation stage to reduce annotation noise while formulating the guidelines for our proposed novel NLP contributions structuring and graphing scheme.

**Findings:** The application of NLPCONTRIBUTIONGRAPH on the 50 articles resulted finally in a dataset of 900 contribution-focused sentences, 4,702 contribution-information-centered phrases, and 2,980 surface-structured triples. The intra-annotation agreement between the first and second stages, in terms of F1-score, was 67.92% for sentences, 41.82% for phrases, and 22.31% for triple statements indicating that with increased granularity of the information, the annotation decision variance is greater.

**Research limitations:** NLPCONTRIBUTIONGRAPH has limited scope for structuring scholarly contributions compared with STEM (Science, Technology, Engineering, and Medicine) scholarly knowledge at large. Further, the annotation scheme in this work is designed by only an intra-annotator consensus—a single annotator first annotated the data to propose the initial scheme, following which, the same annotator reannotated the data to normalize the annotations in an adjudication stage. However, the expected goal of this work is to achieve a standardized retrospective model of capturing NLP contributions from scholarly articles. This would entail a larger initiative of enlisting multiple annotators to accommodate different worldviews into a "single" set of structures and relationships as the final scheme. Given that the initial scheme is first proposed and the complexity of the annotation task in the realistic timeframe, our intra-annotation procedure is well-suited. Nevertheless, the model proposed in this work is presently limited since it does not incorporate multiple annotator worldviews. This is planned as future work to produce a robust model.

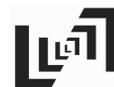

[†] Corresponding author: Jennifer D'Souza (E-mail: jennifer.dsouza@tib.eu).

**Practical implications:** We demonstrate NlpContributionGraph data integrated into the Open Research Knowledge Graph (ORKG), a next-generation KG-based digital library with intelligent computations enabled over structured scholarly knowledge, as a viable aid to assist researchers in their day-to-day tasks.

**Originality/value:** NlpContributionGraph is a novel scheme to annotate research contributions from NLP articles and integrate them in a knowledge graph, which to the best of our knowledge does not exist in the community. Furthermore, our quantitative evaluations over the two-stage annotation tasks offer insights into task difficulty.

**Keywords**    Scholarly knowledge graphs; Open science graphs; Knowledge representation; Natural language processing; Semantic publishing

## 1    Introduction

Our present rapidly amassing wealth of scholarly publications (Jinha, 2010) poses a crucial dilemma for the research community. That is: *how to stay on-track with the past and the current rapid-evolving research progress?* In this era of the publications deluge (Johnson, Watkinson, & Mabe, 2018), such a task is becoming increasingly infeasible even within one's own narrow discipline. The need for novel technological infrastructures in support of intelligent scholarly knowledge access models is, thus, only made more imminent. A viable solution to the dilemma is to make the research progress skimmable for the scientific community with the aid of advanced information access tools. This would help curtail the time-intensive and seemingly unnecessary cognitive labor that currently constitute the researcher's task of searching just for the contribution information in full-text articles to track scholarly progress (Auer, 2018).

Knowledge graphs (KG), i.e. large semantic networks of entities and relationships, are a potent data model solution in this space. KGs enable fine-grained semantic knowledge capture over precise information targets modeled as nodes and links under an optional cumulative knowledge capture theme. Unlike academic articles, KGs facilitate complete programmatic access to their semantified data at the model-specified granularity. Knowledge customizations given targeted queries are possible as knowledge subgraph views over single or aggregated KGs. Consider this in light of the fact, that the researcher's key day-to-day search task over scholarly knowledge is mainly focused on determining scholarly contribution highlights. E.g. "Which is the top-performing system on the SQuAD dataset?" "What methods are used for NER?" etc. KGs are ideal to power such fine-grained knowledge searches. Their well-known utility is in offering enhanced contextualized search as demonstrated successfully in industry by Facebook (Noy et al., 2019) and by Google (A Reintroduction to Our Knowledge Graph and Knowledge Panels, 2020); and even in the open data community by Wikidata (Vrandečić & Krötzsch, 2014) that serves information over many general domains.

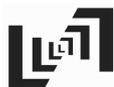

In the technological ecosystem of scholarly KGs, the Open Research Knowledge Graph (ORKG) framework, hosted at TIB[①]—a central library and information center for science and technology—is an emerging initiative geared toward storing scholarly KGs in a digital library of the future (Auer, 2018). It targets the storage of structured scholarly contributions content extensible over the various domains and subdomains in Science at large (Jaradeh et al., 2019). Differing from existing scholarly KG building endeavors that focus on bibliographic and metadata information (Birkle et al., 2020; Hendricks et al., 2020; Wang et al., 2020), the ORKG focuses on organizing the gist of scholarly articles as semantically structured contribution descriptions. It supports a templatized system for the specification of scholarly contributions, thereby obtaining standardized and comparable scholarly knowledge data representations (Oelen et al., 2019; Vogt et al., 2020) which are also in conformance with the FAIR guiding principles for scientific data management and stewardship (Wilkinson et al., 2016).

Given this background, the work described here examines an annotation scheme called the NLPCONTRIBUTIONGRAPH (NCG) (D'Souza & Auer, 2020) that guides the information structuring process of the unstructured scholarly contributions in NLP articles. Application-wise, it specifically targets a scholarly knowledge data acquisition branch of the ORKG, and generally places itself within the wider context of the structuring of scholarly knowledge. By eliciting a set of guidelines for structuring unstructured NLP research contributions, it facilitates instance-based KG data annotation. Figure 1 illustrates a part of an instantiated contributions-focused KG from this scheme.[②] While NCG strives to evolve as a comprehensive templatization model of NLP research contributions, in its present version, it is templated only at the top-level of information organization under the construct of what we refer to as Information Units (IU)—a set of 12 broad information aggregation constructs inspired from scholarly article section names. The rest of the NCG data is modeled based on an annotator's data-driven subjective judgements.

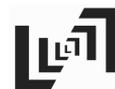

---

[①] https://www.tib.eu/
[②] The full KG for the partial structured contribution information depicted in Figure 1 can be accessed as a resource in the ORKG platform here https://www.orkg.org/orkg/paper/R69764.

Nevertheless, this latter modeling decision subjectivity is to a certain extent, constrained by the IUs. Finally, data annotated in the NCG scheme comprise the following three elements. **1. *Contribution sentences***: a set of sentences about the contribution, where sentences have been adopted in previous scholarly knowledge structuring initiatives albeit with a different semantic focus (Fisas, Ronzano, & Saggion, 2016; Liakata et al., 2010; Teufel, Carletta, & Moens, 1999; Teufel, Siddharthan, & Batchelor, 2009). **2. *Scientific terms and relations***: a set of scientific term and predicate phrases extracted from the contribution sentences, where scientific terms and predicates have also been considered in prior knowledge structuring annotation initiatives (Augenstein et al., 2017; Luan et al., 2018). **3. *Triples***: semantic statements of related subject-object scientific term pairs toward RDF statement representations for KG building. The complete RDFization of NCG triples would entail an additional linking of the scientific terms and predicates to existing ontology resource URIs by synonymy identification and ontology alignment. An alternative and simpler strategy would be the RDFization of the NCG triples by simply coining own URIs. However, the latter method would create large amounts of new resources rather than leveraging existing ones.

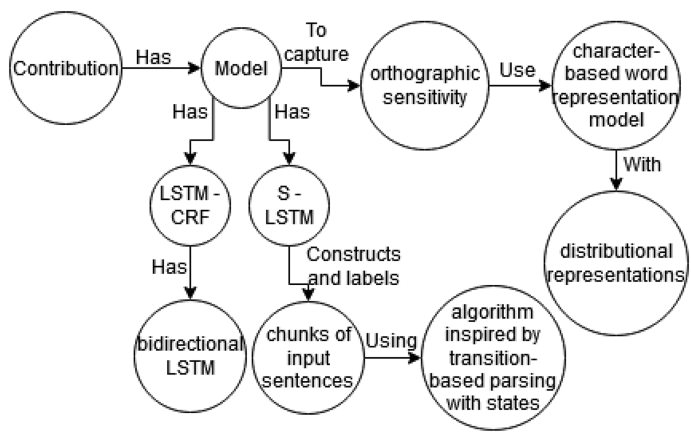

Figure 1.   Structured Model information as part of the research contribution highlights of a scholarly article (Lample et al., 2016) in the NLPCONTRIBUTIONGRAPH scheme.

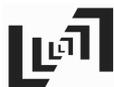

In contrast to other existing content-based scholarly KG generation methods (Buscaldi et al., 2019; Jiang et al., 2020; Luan et al., 2018), NCG has an overarching knowledge capture theme, i.e. *to capture only the scholarly articles' "contributions."* With this overarching knowledge capture objective, the first version of NCG (2020) was developed over the course of an annotation exercise performed on 50 NLP articles which were uniformly selected across five different NLP subfields: 1. machine translation, 2. named entity recognition, 3. question answering, 4. relation classification, and 5. text classification.

In this study, the NCG scheme is revisited with a two-fold objective: 1) to identify any redundancies in the representation and thereby normalize them; and 2) to obtain a fairly reliable and consistent set of annotation guidelines. Building on our prior work, in this article, we re-annotated the same set of 50 articles a second time and examined the changes obtained via this adjudication task. Specifically, the following questions were investigated:

- How data intensive is the annotation procedure?—i.e. what proportion of the full-text article content constitutes core contribution information, and consequently, the structured data within this scheme?
- Were significant changes needed to be made to the annotation scheme between the pilot and the adjudication phases?—i.e. were large quantified changes observed in the intra-annotation measures?

Summarily, NCG informs instance-based KG generation over NLP scholarly articles where the modeling process is mostly data-driven, unguided by a specific ontology, except at the top-level categorization of the information under IUs. Nevertheless, a large dataset of annotated instances by the NCG scheme would be amenable to ontology learning (Cimiano et al., 2009) and concept discovery (Lin & Pantel, 2002). The NCG data characteristically caters to practical applications such as the ORKG (Jaradeh et al., 2019) and other similar scholarly KG content representation frameworks designed for the discoverability of research conceptual artefacts and comparability of these artefacts across publications (Oelen et al., 2020) which we demonstrate concretely in Section 6. By adhering to data creation standards, the NCG by-product data when linked to web resources will fully conform to the FAIRness principle for scientific data (Wilkinson et al., 2016) as its data elements will become Findable, Accessible, Interoperable, and Reusable.

## 2 Related work

Early initiatives in semantically structuring scholarly articles focused just on sentences as the basic unit of analysis. To this end, ontologies and vocabularies were created (Constantin et al., 2016; Pertsas & Constantopoulos, 2017; Soldatova & King, 2006; Teufel et al., 1999), corpora were annotated (Fisas et al., 2016; Liakata et al., 2010), and machine learning methods were applied (Liakata et al., 2012). Like NLPCONTRIBUTIONGRAPH, these prior sentence annotations were thematically-based, ranging from fine-grained rhetorical themes modeled after scholarly article sections (Fisas et al., 2016; Liakata et al., 2010; Teufel, Carletta, & Moens, 1999; Teufel, Siddharthan, & Batchelor, 2009), to identifying scientific knowledge claim arguments (Teufel, Siddharthan, & Batchelor, 2009). Where the first kind of structuring practically enabled highlighting sentences with fine-grained semantics within computer-based readers, the latter served citation summaries and sentiment analysis.

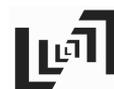

Following sentence-based annotations, the ensuing trend for structuring the scholarly record has the specific aim of bolstering search technology. Thus, it was steered towards scientific terminology mining and keyphrase extraction. Consequently, this led to the release of phrase-based annotated datasets in various domains including multidisciplinarily across STEM (Augenstein et al., 2017; D'Souza & Auer, 2020; Handschuh & QasemiZadeh, 2014; Luan et al., 2018), which facilitated machine learning system development for the automatic identification of scientific terms from scholarly articles (Ammar et al., 2017; Beltagy, Lo, & Cohan, 2019; Brack et al., 2020; Luan, Ostendorf, & Hajishirzi, 2017).

While NLPCONTRIBUTIONGRAPH incorporates sentence and phrasal-unit data elements, it differs from all prior attempts to structure or semantify scholarly articles. To the best of our knowledge, no prior model has attempted gathering data elements over the theme of scholarly contributions. Closer to our work of adopting a thematic focus in practically leveraging knowledge graphs, increasingly, text mining initiatives are seeking out to structure recipes or formulaic semantic patterns as KGs (Kononova et al., 2019; Kulkarni et al., 2018; Kuniyoshi et al., 2020; Mysore et al., 2019). These efforts advocate for machine-interpretable documentation formats of wet lab protocols and inorganic materials synthesis reactions and procedures for faster computer-mediated analysis and predictions. Relatedly, the objective of NCG is to obtain machine-interpretable NLP scholarly contributions to foster the automatic generation of surveys.

## 3 The NLPCONTRIBUTIONGRAPH scheme: Preliminaries

The NCG scheme aims to build a scholarly KG assuming a bottom-up data-driven design. Thus, while not a fully ontologized model, it has one top-level layer predefined with a set of content category types for surface typing of the contribution knowledge that the graph represents. This follows the idea of content organization as scholarly article sections. However, in the NCG, the types are a predefined closed class set similar to the introduction, methods, results, and discussion (IMRAD) format prescribed for medical scientific writing (Huth, 1987; Sollaci & Pereira, 2004). Next, we describe this knowledge systematization.

The NCG scheme has two levels of content typing as follows: 1) it has a root node called "Contribution." By means of this node, an instantiated NCG can be attached to another KG. E.g. extending the ORKG by attaching an instantiated NCG's "Contribution" node to it. 2) In the next level, it has 12 nodes referred to as IU.

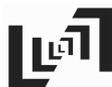

Each scholarly article's annotated contribution data elements are organized under three (mandatory) or more of the IU nodes. These nodes are briefly described next.

## 3.1 Twelve information unit nodes

Annotating structured contributions information from scholarly articles (see Section 4 for details), showed that, per article, its contribution-centered content could be organized as three or more different rhetorical categories inspired from scholarly article section names. Specifically, 12 different contribution content types were identified, a few of which were in common with the fine-grained rhetorical semantic classes annotations made for scholarly article sentences (Teufel, Siddharthan, & Batchelor, 2009). The 12 types are as follows.

**i. RESEARCHPROBLEM** determines the research challenge addressed by a contribution using the predicate *has Research Problem*. By definition, it is the focus of the research investigation; in other words, *the issue for which the solution must be obtained*. It is typically found in an article's Title, Abstract, and in the first few paragraphs of the Introduction.

**ii. APPROACH** or **iii. MODEL** This is the contribution of the paper as *the solution proposed for the research problem*. The choice of the name, i.e. whether it is APPROACH or MODEL, depends on the paper's content. As a guiding rule of thumb, it is called APPROACH when the solution is proposed as an abstraction, and is called MODEL if the solution is proposed in practical implementation terms. Further, in the paper, the solution may not always be introduced as *approach* or *model*; in which case, the names have to be normalized to either APPROACH or MODEL. E.g. if the solution is introduced as "method" or "application," this is normalized as APPROACH; on the other hand, if it is referred to as "system" or "architecture," it is normalized to MODEL. In terms of the content captured, only the highlights of the solution are relevant in our contribution-focused model—in-depth details such as architectural figures or equations are not included. This information is found in the article's Introduction section in the context of cue phrases such as "we take the *approach*," "we propose the *model*," "our system *architecture*," or "the *method* proposed in this paper." However, there are exceptions when the Introduction does not present an overview of the system, in which case we analyze the first few lines of the system description section itself in the article. The APPROACH or MODEL is connected to the Contribution root node via the predicate *has*—this is followed for the remaining nine units as well.

**iv. CODE** is a link to the software on open-source hosting platforms such as Github or Gitlab, or even on the author's website. This information, when present, can generally be found in the Abstract or Introduction, and occasionally in the Methods or Conclusion sections.

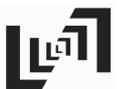

**v. DATASET** is sometimes included as a contribution information unit when it is a solution proposed in an NLP article. E.g. the SQuAD dataset that defines a Question Answering task.

**vi. EXPERIMENTALSETUP** or **vii. HYPERPARAMETERS** include hardware (e.g. GPU) and software (e.g. Tensorflow③, Weka④) details for implementing the machine learning solution; and variables that determine the network structure (e.g. number of hidden units) and how the machine learning model is optimized (e.g. learning rate, regularization) to the task objective. When details of the hardware are also given, it is called EXPERIMENTALSETUP; otherwise, HYPERPARAMETERS. This IU offers users better insights of the machine learning model. Further, it is only modeled when expressed in simple sentence constructs. When they have complex descriptions, it is not easily possible to structure the data. E.g. "machine translation" models which tend to be fairly complex with many parameters. Thus, the decision to model the experimental setup, seemingly a subjective decision at face value, over the course of several annotated articles, becomes apparent with better familiarity of the eligible sentence patterns.

**viii. BASELINES** are systems that the proposed APPROACH or MODEL is compared against. It structures information about which existing systems the proposed model is compared to.

**ix. RESULTS** are *the main findings or outcomes reported in the article* for the RESEARCHPROBLEM. This content is generally found toward the end of the article in the RESULTS, EXPERIMENTS, or TASKS sections. While the results are often highlighted in the Introduction, unlike the APPROACH or MODEL units, in this case, we annotate the dedicated, detailed section on RESULTS because results constitute a primary aspect of the contribution.

Next, we discuss the 10$^{th}$ IU called TASKS which can either be encapsulated in the RESULTS IU or which encapsulates several RESULTS IU against different tasks.

**x. TASKS** in multi-task settings, i.e. when the solution is tested on more than one task, list all the experimented tasks. Similarly, if experimental datasets are listed, they can be interpreted as tasks since it is common in NLP for tasks to be defined over datasets. Each task listed in TASKS can include one or more of the EXPERIMENTALSETUP, HYPERPARAMETERS, and RESULTS as sub-information units.

**xi. EXPERIMENTS** is an encapsulating IU with one or more of the above five results-focused IUs. It can include a combination of EXPERIMENTALSETUP, lists of TASKS and their RESULTS, or can be a combination of APPROACH, EXPERIMENTALSETUP and RESULTS. It resembles the scholarly article sections "experiment" or

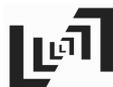

③  https://www.tensorflow.org/
④  https://www.cs.waikato.ac.nz/ml/weka/

"experiments." Recently, more and more multitask systems are being developed, e.g. BERT (Devlin et al., 2018). Therefore, modeling EXPERIMENTALSETUP with TASKS and RESULTS is necessary for such systems since the experimental setup often changes per task producing a different set of results. Hence, this information unit encompassing two or more sub information units is part of the NCG scheme.

**xii. ABLATIONANALYSIS** is a type of RESULTS that shows in fine-grained detail the contribution of each component in multi-component systems proposed for NLP tasks as the APPROACH or MODEL. In some articles, an ablation study is conducted to evaluate the performance of a system by removing certain components, to understand the contribution of the component to the overall system. Therefore, this IU is used to model these results if they are expressed in a few sentences, similar to our modeling criteria for HYPERPARAMETERS.

This concludes the detailed description about the 12 top-level IU nodes in the NCG scheme. Of the 12, only three are mandatory for structuring contributions per scholarly article. They are: RESEARCHPROBLEM, APPROACH or MODEL, and RESULT— the remaining may or may not be present based on the content of the article.

## 4 The NLPCONTRIBUTIONGRAPH scheme: Annotation exercise

### 4.1 Stage 1: Pilot annotations

The pilot annotation task (D'Souza & Auer, 2020) that resulted in the preliminary version of the NCG scheme was performed on a set of 50 NLP scholarly articles. This set of 50 articles constituted the trial dataset. Thus, the preliminary NCG scheme was defined over the trial dataset in a data-driven manner during the pilot annotation exercise.

There were two requirements decided at the outset of the annotation task. First and foremost, the graph model needed to be a *robust* representation for the diversity in NLP scholarly contributions. To facilitate this, a representative dataset of the NLP domain was needed so that unique aspects of *contributions* could be observed. Thus, the articles were uniformly chosen across five different NLP subfields: 1. machine translation, 2. named entity extraction, 3. question answering, 4. relation classification, and 5. text classification.[⑤]

The second design choice, regards the granularity of the data annotated in the NCG scheme. In the Related Work (Section 2), we saw that the sentential and phrasal granularity was used in prior work on structuring scholarly articles. Thus, inspired from this prior annotation science and also toward modeling KGs, the

---

[⑤] These five tasks were randomly selected among the most popular NLP tasks on the paperswithcode.com leaderboard.

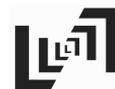

following three different granularity levels were established. At the first level, *sentences* that capture information about an article's contribution; at the second level, *phrases* about scientific knowledge terms and relation predicates identified from the selected contribution-information sentences; and, at the third level, toward building KGs, *triples* uniting the scientific terms and predicates as KG instances.

Table 1 below illustrates two examples of modeling contribution-oriented sentences, phrases, and triples from a scholarly publications categorized under the RESULTS IU. Examples 1a and 2a, respectively, first show the sentences information granularity in the NCG scheme. These sentences discuss performance comparisons of the proposed model with existing scores. In general, for the RESULTS IU, similar sentences that discuss performance comparisons of the proposed system versus existing systems should be selected as NCG model instances. Next, as examples of the *phrasal granularity*, 1b and 2b reflect the scientific knowledge terms and predicates from the sentences 1a and 2a, respectively. These are selected such that they should be able to take one of three roles, i.e. subject, predicate, or object, in triples for building KGs. Linguistically, they span a diverse spectrum of types, i.e. they can be noun phrases or verb phrases or adjectival phrases, etc. Further, the types are not related to the triple roles. Consider from 1b, the verb phrase "adding features" or the noun phrase "neural networks" can be either subject or object, whereas the phrase "computed by" which is the beginning part of a verb phrase is a predicate indicating a relation. Last, 1c and 2c illustrates the arrangement of the phrases as subject-predicate-object KG triples. Note, however, the order of the annotation steps need not be the same as the examples. At the least, we imagine sentences necessarily are first in the annotation task order. Following which, the annotator can perform the triples annotation first and later break their structure into phrases or the reverse. In the reverse case, they would make implicit triples considerations. In this context, one might wonder why include the *phrases* data element at all as part of the NCG scheme. We do this to enable flexibility in problem formulation for machine learning and, more concretely, to enable a clearer evaluation scenario since it could be that a machine learner may not be very accurate in forming triples but is better at identifying phrases.

We refer the reader to our prior work (D'Souza & Auer, 2020) for additional details regarding the pilot annotation task itself.

### 4.2  Stage 2: Adjudication annotations

We carried out a two-stage annotation cycle over the trial dataset to finalize the NCG scheme. The first stage was the *pilot* annotation cycle with a brief overview provided in the earlier section (Section 4.1). The second stage was the *adjudication* cycle during which the graph model was finalized and the gold-standard article-wise

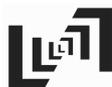

Table 1.   Two examples illustrating the three different granularities for NlpContributionGraph data instances (viz., a. sentences, b. phrases, and c. triples) modeled for the Result information unit from a scholarly article (Cho et al., 2014).

[**1a.** *sentence 159*] As expected, adding features computed by neural networks consistently improves the performance over the baseline performance.
[**1b.** *phrases from sentence 159*] {adding features, computed by, neural networks, improves the performance, over baseline performance}
[**1c.** *triples from entities above*] {(Contribution, has, Results), (Results, improves the performance, adding features), (adding features, computed by, neural networks), (Results, improves the performance, over baseline performance)}

[**2a.** *sentence 160*] The best performance was achieved when we used both CSLM and the phrase scores from the RNN Encoder – Decoder.
[**2b.** *phrases from sentence 160*] {best performance was achieved, used both CSLM and the phrase scores, from, RNN Encoder – Decoder}
[**2c.** *triples from entities above*] {(Contribution, has, Results), (Results, best performance was achieved, used both CSLM and the phrase scores), (used both CSLM and the phrase scores, from, RNN Encoder – Decoder)} graph data were obtained. A gap of two weeks separated the *pilot* and *adjudication* stages. Since the model development and annotation exercise was performed by a single annotator—an NLP domain specialist—, the stages needed to be sufficiently apart in time to control for the decision-making thought process from the pilot stage influencing adjudication. Nonetheless, the second stage was mere adjudication, therefore was non-blind. This meant that the annotator has access to the pilot annotations when re-annotating the data. Our adjudication objective was not to arrive at a new model that significantly differed from the model in the pilot stage, but it was to normalize that model itself. In other words, by adjudications, we merely sought to refine the quality of the scheme. Given the complex nature of this annotation task (cf. annotation evaluations in Section 5), and the need to design a model within a realistic timeframe, our annotation procedure is well-suited. Conversely, adopting a different annotation procedure, i.e. perhaps a blind second stage, might have resulted in a model with no consensus at all, since a consensus would have taken an unrealistic amount of time to arrive at—consider that since the NCG scheme is designed over the full-text of the scholarly articles its data consideration scope is wide despite the information constraint of the IUs. Further, using a different annotator for the adjudication stage than in the pilot stage could potentially result in a new model if the two annotators had different interpretations of contribution information—neither being incorrect, just different. This is a common phenomenon observed in the ontology engineering community where several ontologies exist under one knowledge theme but still modeling different worldviews. We also advocate for a blind, multi-stage, and multi-annotator annotation process for the NCG scheme, recognizing it as a potentially better annotation model since it can produce a well-defined set of annotation rules based on strict linguistic patterns agreed upon by multiple views. However, this is relegated as future work with a stipulated time-frame of at least two years. On the other hand, our consensus-oriented method is practically better suited as a first step toward designing complex graph-based models, in other words, to obtain the NCG scheme in a realistic timeframe.

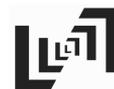

There were two requirements decided at the outset of the adjudication annotation task. They were: 1) to normalize IUs further to be a smaller, but comprehensively representative, set of similar structured properties to facilitate succinct contribution comparisons across articles' contribution graphs; and 2) to improve the phrasal boundary decisions made in the pilot stage focused on targeting precise scientific knowledge semantics within the annotated phrases. Otherwise, both the pilot and adjudication stages adopted the same annotation workflow as depicted in Figure 2.

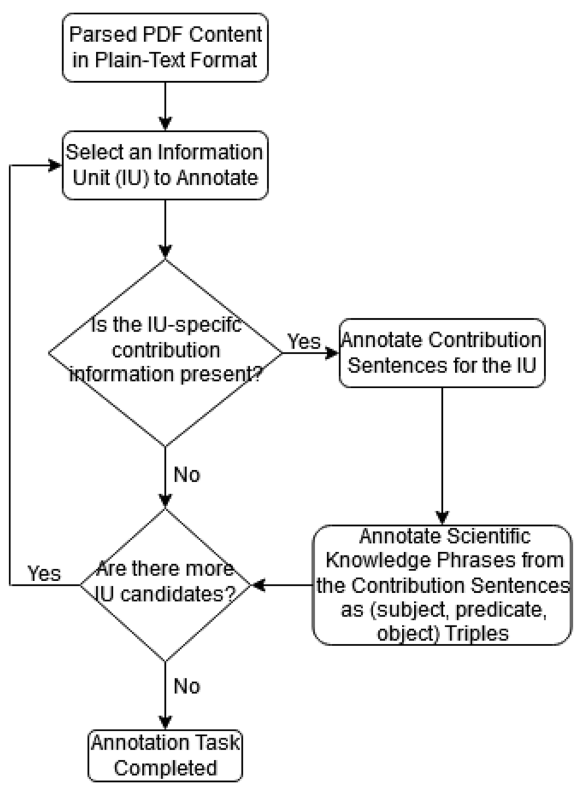

Figure 2.   Functional workflow of the annotation process to obtain the NLPCONTRIBUTIONGRAPH data.

Let us elaborate on the two requirements: 1) normalizing IUs—we had 16 different IUs in the pilot stage. During the course of the adjudication stage annotations, they were normalized as the following nine IUs: RESEARCHPROBLEM, EXPERIMENTALSETUP,

HYPERPARAMETERS, RESULTS, TASKS, EXPERIMENTS, ABLATIONANALYSIS, BASELINES, and CODE were retained as is; the set MODEL, APPROACH, METHOD, ARCHITECTURE, SYSTEM, and APPLICATION was reduced to MODEL and APPROACH—in our second round of observations of the data, we identified references to METHOD, ARCHITECTURE, SYSTEM, and APPLICATION all pointed at an actual piece of software rather than a conceptual solution, thus they were normalized into a single MODEL IU, while APPROACH referred to the conceptual solution; and the OBJECTIVE IU was dropped since it was infrequent. Further, a DATASET IU was added, since in NLP papers they are sometimes described as a solution. 2) improving phrasal boundaries—this involved making smaller split decisions on phrases to convert very specific scientific knowledge elements into entities that could take on more generic roles and therefore were reusable in a KG. E.g. "models that use extensive sets of handcrafted features" split into the following three phrases "models," "use," and "extensive sets of handcrafted features." This example is depicted in detail in Appendix 2.

### 4.2.1 The NCG Scheme's five general annotation guidelines

Finally, five main annotation guidelines are prescribed for NCG.

1) Sentences with contribution data are identified in various places in the paper including title, abstract, and full-text. Within the full-text, only the Introduction and Results sections are annotated. Sometimes, the first few sentences in the Methods section are annotated as well if method highlights are unspecified in the Introduction.
2) Only sentences that directly state the paper's contribution are annotated.
3) All relation predicates are annotated from the paper's text, except the following three, i.e. *has*, *name*, and *hasAcronym*. "*has*" is used for connecting the IU nodes to the Contribution root node, and as a filler predicate if one is not found directly in the text. "*name*" is the filler predicate used to link the model name to the APPROACH or MODEL nodes. "*hasAcronym*" has a similar function to *name*, except applied only for acronyms of names.
4) Past the IU level, for parent node names in the graph, the names of sections are preferred, which if challenging to annotate are identified in the running text (see Appendix 1 example).
5) Repetitions of the scientific terms or predicates, which do not correspond to the actual information in the text, are not allowed when forming KG triples. See illustrated in Figure 3.

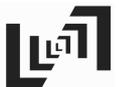

```
{
  "has" : {
    "Results" : {
      "in terms of" : {
        "F1 measure" : {
          "in" : {
            "ACE datasets" : {
              "achieves" : "best results"
            },
            "GENIA dataset" : {
              "achieves" : "comparable results"
            }
          }
        }
      },
      "from sentence" : "Our neural transition
        -based model achieves the best results in
        ACE datasets and comparable results in
        GENIA dataset in terms of F1 measure."
    }
```

Figure 3.   Illustration of the annotation guideline 5 of forming triples without incorrect repetitions of the extracted phrases. This RESULTS IU is modeled from the research paper by (Wang et al., 2018). If the phrases "in terms of" and "F1 measure" were modeled by sentence word order, they would need to be reused twice under the "ACE datasets" and "GENIA dataset" scientific terms. To avoid this incorrect repetition, despite being at the end of the sentence, they are annotated at the top of the triples hierarchy.

## 5  The NLPCONTRIBUTIONGRAPH Scheme: Evaluating the annotations

### 5.1  Raw data and preprocessing tools

The trial dataset for designing the NCG scheme was derived from a collection downloaded from the publicly available leaderboard of tasks in artificial intelligence called https://paperswithcode.com/. The paperswithcode.com dataset predominantly contained articles in the Natural Language Processing and Computer Vision domains. To design the NCG scheme, we restricted ourselves just to its NLP papers. From the original collection of papers, we randomly selected 10 papers from each of the five different NLP subfields which resulted in a total of 50 papers. The five NLP subfields were: 1. machine translation, 2. named entity recognition, 3. question answering, 4. relation classification, and 5. text classification.

The raw data needed to undergo a two-step preprocessing to be ready for analysis. 1) For pdf-to-text conversion of the scholarly articles, the GROBID parser (GROBID, 2008) was applied; following which, 2) for plaintext pre-processing in terms of tokenization and sentence splitting, the Stanza toolkit (Qi et al., 2020) was used. The resulting pre-processed articles were then leveraged in the two-stage annotation cycle (see Section 4) to devise the NCG scheme.

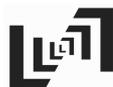

## 5.2 Annotated corpus statistics

The overall annotated corpus statistics for our trial dataset after the adjudication stage is depicted in Table 2. We see that, in each of the five subfields, approx. 40 IUs were annotated on each of the subfield's articles, i.e. on average four IUs per article per subfield since 10 articles were selected for each subfield. This implies that, on average, each article was annotated with more or less one additional IU beside the three (RESEARCHPROBLEM, APPROACH or MODEL, and RESULTS) deemed mandatory. Further, in the annotated data, per subfield, we see at the sentence element level, that a very small fraction (< 0.1%) of the scholarly article sentences constituted contribution sentences (see Table row "*avg ann Sentences*"). In terms of tokens, < 0.3% of the overall tokens were annotated as scientific terms or predicates toward KG building (see Table row "*avg ann Phrase Toks*"). Finally, in terms of triples, approx. 600 triples were annotated per task, i.e. nearly 60 triples per article per task. Thus, overall, while not much data is annotated per article (approx. 16 sentences and 90 phrases per article—see Table rows "*ann Sentences*" and "*ann Phrases*" for the total counts), the task challenge is selecting the pertinent contribution-focused information given the wide range of the available candidate data.

Next we highlight a few differences between the five subfields in our dataset. In Table 2, we see that machine translation (MT) had the most annotated sentences. This can be attributed to the observation that generally MT articles tend to be longer descriptively and particularly in terms of models settings compared to the other four subfields. Further, relation classification (RC) had the highest proportion of contribution sentences constituting its articles. With 0.1 this still indicates a low proportion reflecting the fact that contribution information is contained in relatively few sentences. Text classification (TC) had the highest number of annotated phrases despite not being among the tasks with the highest numbers of annotated sentences. This implies that the number of annotated sentences is not directly related to the number of annotated phrases for the tasks in our data. But this understandably is not the same concerning the correlation between the number of phrases and triples, wherein the number of phrases and triples are directly related.

Table 3 depicts the final annotated corpus statistics in terms of the information units. We make the following key observations: RESULTS has the most number of triples; RESEARCHPROBLEM is modeled for all articles; and EXPERIMENTS has the highest triples-to-papers ratio. We see that EXPERIMENTS is annotated only in three articles yet contains a high number of triples. This can be attributed to the fact that this particular IU can also encapsulate other IUs such as EXPERIMENTALSETUP, TASKS, and RESULTS (cf. Section 3.1). This is different from the other IUs that have no other IU encapsulation, with TASKS as an exception.

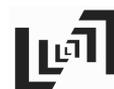

Table 2. Annotated corpus characteristics for our trial dataset containing a total of 50 NLP articles using the NLPCONTRIBUTIONGRAPH model. "ann" stands for annotated; and IU for information unit. The 50 articles are uniformly distributed across five different NLP subfields characterized at sentence and token-level granularities as follows—machine translation (MT): 2,596 total sentences, 9,581 total overall tokens; named entity recognition (NER): 2,295 sentences, 8,703 overall tokens; question answering (QA): 2,511 sentences, 10,305 overall tokens; relation classification (RC): 1,937 sentences, 10,020 overall tokens; text classification (TC): 2,071 sentences, 8,345 overall tokens.

|                      | MT    | NER   | QA    | RC    | TC    | Overall |
|----------------------|-------|-------|-------|-------|-------|---------|
| total IUs            | 38    | 43    | 44    | 45    | **46**| 216     |
| ann Sentences        | **209**| 157  | 176   | 194   | 164   | 900     |
| avg ann Sentences    | 0.081 | 0.068 | 0.07  | **0.1**| 0.079| -       |
| ann Phrases          | 956   | 770   | 960   | 978   | **1038**| 4,702 |
| avg Toks per Phrase  | 2.81  | 2.87  | 2.76  | **2.91**| 2.7 | -       |
| avg ann Phrase Toks  | 0.28  | 0.25  | 0.26  | 0.28  | **0.34**| -     |
| ann Triples          | 590   | 504   | 619   | 620   | **647**| 2,980  |

Table 3. Annotated corpus statistics for the 12 Information Units in the NLPCONTRIBUTIONGRAPH scheme.

| Information Unit | No. of triples | No. of papers | Ratio of triples to papers |
|---|---|---|---|
| EXPERIMENTS | 168 | 3 | **56** |
| TASKS | 277 | 8 | 34.63 |
| EXPERIMENTALSETUP | 300 | 16 | 18.75 |
| MODEL | 561 | 32 | 17.53 |
| HYPERPARAMETERS | 254 | 15 | 16.93 |
| RESULTS | **688** | 42 | 16.38 |
| APPROACH | 283 | 18 | 15.72 |
| BASELINES | 148 | 10 | 14.8 |
| ABLATIONANALYSIS | 155 | 13 | 11.92 |
| DATASET | 8 | 1 | 8 |
| RESEARCHPROBLEM | 169 | **50** | 3.38 |
| CODE | 9 | 9 | 1 |

### 5.3 Intra-Annotation agreement measures

We now compute the intra-annotation agreement measures between the first and the second stage versions of the dataset annotations across all three data elements in the NCG scheme including its top-level information units. Our evaluation metrics are the standard precision, recall, and F1-score.

Table 4 depicts the results. With these scores, we quantitatively observe the degree of changes between the two annotation stages treating the second stage as reference gold-standard. Between the two stages, the F1-scores of the annotation changes were: information units 79.64%, sentences 67.92%, phrases 41.82%, and triples 22.31%. We conclude that the interpretation of annotations related to the top-level organization of scholarly contributions did not change significantly (at 76.64% F1-score). Even the decision of the annotator about the sentences containing contribution-centered information showed a low degree of change (at 67.92%

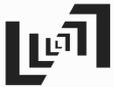

F1-score). However, the comparison of the fine-grained organization of the contribution-focused information, as phrases or triples, obtained low F1-socres. Finally, from the results, we see that our pipelined annotation task is presented with the general disadvantage of pipelined systems, wherein the performances in later annotation stages is limited by performances in earlier annotation stages.

In light of the low intra-annotator agreement obtained at the phrasal and triples information granularities, a natural question may arise: *is our proposed model valid at all?* We claim validity on the following basis. 1) Since the data was annotated by a single person, there is a data modeling uniformity in their decisions arising from the fact that it was made by the same person. Thus, the annotated data in any stage consistently models the annotator decisions in that stage. For instance, regarding the sentences, major discrepancy arose between the pilot and the adjudication stages from decisions such as selecting titles as contribution sentence candidates, wherein in the first annotation stage, titles were not annotated, but were uniformly selected to be annotated in the adjudication stage. 2) Further, as shown in the "*avg ann Sentences*" row in Table 2, since roughly only 8% of the article sentences were contribution sentences, an expected consequence of this task is indeed a high amount of variance in the sentence selection decisions. Such a discrepancy can only be resolved with further annotation stages. With reference to our earlier example, since titles were uniformly established as contribution sentence candidates in the second annotation round, in a third round we would have much higher modeling decision scores. 3) Finally, our annotation task has a pipelined nature. Thus, sentence modeling discrepancies are magnified in the later annotation stages, i.e. for phrases or triples annotations. This explains a major proportion of the low agreement scores in the second annotation round for phrases and triples since new phrases (and triples) had to be annotated for the new added sentences.

Observing the task-specific intra-annotation measures in rows 1 to 5, we find that question answering (QA) and text classification (TC) have the highest scores reflecting fewer changes made in their annotations than the other tasks, albeit at decreasing levels across the data elements.

Generally, one may wonder why triples formation is challenging given a set of scientific term and predicate phrases indicated by its lowest F1-score per task and overall. There are two reasons: a) the phrases were significantly changed in the second stage (see Phrases F1-scores as < 50%), which in turn impacted the triple formation; and b) often the triples can be formed in more than one way. In the adjudication process, it was established for the triples to strictly conform to order of appearance of phrases in the text, which in the first stage was not consistently followed.

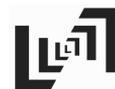

Table 4. Intra-Annotation Evaluation Results. The NLPCONTRIBUTIONGRAPH scheme pilot stage annotations evaluated against the adjudicated gold-standard annotations made on the trial dataset.

|  | Tasks | Information Units | | | Sentences | | | Phrases | | | Triples | | |
|---|---|---|---|---|---|---|---|---|---|---|---|---|---|
|  |  | P | R | F1 | P | R | F1 | P | R | F1 | P | R | F1 |
| 1 | **MT** | 66.66 | 73.68 | 70.0 | 66.67 | 54.55 | 60.0 | 37.47 | 30.96 | 33.91 | 19.73 | 17.46 | 18.53 |
| 2 | **NER** | 79.55 | 81.40 | 80.46 | 60.89 | 69.43 | 64.88 | 44.09 | 42.60 | 43.34 | 22.34 | 21.63 | 21.98 |
| 3 | **QA** | 93.18 | 93.18 | **93.18** | 67.96 | 79.55 | 73.30 | 54.04 | 45.21 | 49.23 | 37.50 | 32.0 | **34.52** |
| 4 | **RC** | 70.21 | 73.33 | 71.74 | 64.64 | 60.31 | 62.40 | 35.31 | 29.24 | 32.0 | 12.59 | 11.45 | 11.99 |
| 5 | **TC** | 86.67 | 84.78 | 85.71 | 75.44 | 78.66 | **77.01** | 54.77 | 45.38 | **49.63** | 27.41 | 22.41 | 24.66 |
| *Cum.* | *micro* | 78.83 | 80.65 | 79.73 | 67.25 | 67.63 | 67.44 | 45.36 | 38.83 | 41.84 | 23.76 | 20.97 | 22.28 |
|  | *macro* | 78.8 | 80.49 | 79.64 | 67.33 | 68.51 | 67.92 | 45.2 | 38.91 | 41.82 | 23.87 | 20.95 | 22.31 |

## 6  The NLPCONTRIBUTIONGRAPH Scheme: Practical use case

The NCG scheme was designed to structure NLP contributions thereby generating a contributions-centric KG. Such data will ease the information processing load for researchers who presently invest a large share of their time in surveying their field by reading full-text articles. The rationale for designing such a scheme was the availability of scholarly KGs, such as the ORKG (Jaradeh et al., 2019), that are equipped with features to automatically generate tabulated comparisons of various approaches addressing a certain research problem on their common properties and values (Oelen et al., 2020). We integrate some of our articles' structured contributions into the Open Research Knowledge Graph (ORKG). Tapping into the ORKG's contributions comparison generator feature over our structured data, we demonstrate how *automatically generated tabulated surveys* can be obtained given scholarly contribution KGs. Such an information processing tool can easily assist the researcher in their day-to-day task of keeping track of research progress, potentially reducing their cognitive effort expended from weeks or months to a matter of just minutes.

In the following subsections, we first describe how an article's contribution data modeled by NCG is integrated in the ORKG, and then illustrate the comparison feature.

### 6.1  Leveraging the Open Research Knowledge Graph framework

The ORKG comprises structured descriptions of research contributions per article. The user can enter the relevant data about their papers via the framework online at https://orkg.org. In Figure 4 below, we depict our annotated triples in plain text format for the RESULTS IU capturing just the results contribution aspect of a paper. Note that this data can alternatively be represented in JSON format (as in Figure 3 shown earlier in this paper).

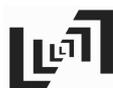

```
(Contribution||has||Results)
(Results||on||QASent dataset)
(QASent dataset||got||comparable MRR)
(comparable MRR||than||dos)
(QASent dataset||got||best MAP)
(best MAP||among||all previous work)
(Results||on||MSRP dataset)
(MSRP dataset||obtained||comparable performance)
(comparable performance||without using||any sparse features)
(comparable performance||without using||extra annotated resources)
(comparable performance||without using||specific training strategies)
(Results||on||Wiki QA dataset)
(Wiki QA dataset||more effective than||other models)
```

Figure 4. Annotated data from the paper "Sentence similarity learning by lexical decomposition and composition" under the RESULTS Information Unit by the NLPCONTRIBUTIONGRAPH scheme.

With the help of the ORKG paper editor interface, we add a paper to the ORKG including its bibliographic information, the RESULTS IU data (Figure 4), and the RESEARCHPROBLEM IU data. This paper view in the ORKG is depicted in Figure 5 below. In the figure, while the data values for RESEARCHPROBLEM IU in the NCG scheme is directly visible, the data for the RESULTS is not. This is due to the paper-view showing only the top-level node. To access deeper levels of the structured data, one would need to click on the Results text link in orange. Such a branch traversal starting at Results until the last node is depicted in Figure 6 as a four-part series.

With this we have described how the structured contributions data of individual papers are represented in the ORKG. Next, we showcase the ORKG feature for generating comparisons.

## 6.2 Automated NLP contribution comparisons

The ORKG has a feature to generate and publish surveys in the form of tabulated comparisons over articles' knowledge graph nodes (Oelen et al., 2020). To demonstrate this feature, we entered our data for the RESULTS IUs of four papers including the one depicted in Figure 5 in the ORKG. Applying then the ORKG survey feature on the four structured articles contributions aggregates their semantified graph nodes in a tabulated comparison across them. This is depicted in Figure 7. This computation aligns closely with the notion of traditional survey articles, except it is automated and operates on machine-actionable knowledge elements. With such an aggregate view over common contribution properties it is easier for users to ingest the essential details of the articles in a matter of minutes in a single view, a feature that is even more advantageous when a larger number of articles is compared. Given that the data is directly extracted from the text it is amenable to further processing for better knowledge representation quality (e.g. "Outperforming," "Outperforms," and "Significantly outperforms" can be normalized as a single predicate).

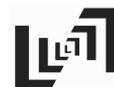

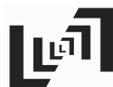

Figure 5. An Open Research Knowledge Graph paper view. The NLPCONTRIBUTIONGRAPH scheme is employed to model the RESEARCHPROBLEM and the RESULTS information units of the paper.

Thus we have demonstrated how structured contributions from the NCG scheme address the massive scholarly knowledge content ingestion problem.

## 8  Conclusions and future directions

We have discussed the NCG scheme for structuring research contributions in NLP articles as structured KGs. We have described the process of leveraging the NCG scheme to annotate contributions in our trial dataset of 50 NLP articles in two stages which helped us obtain the NCG annotation guidelines and improve data quality. Further, we demonstrated how such structured data is poignant in the face of growing volumes of scholarly content to help alleviate the scholarly knowledge content ingestion problem. Our annotated dataset is publicly available here: https://github.com/ncg-task/trial-data.

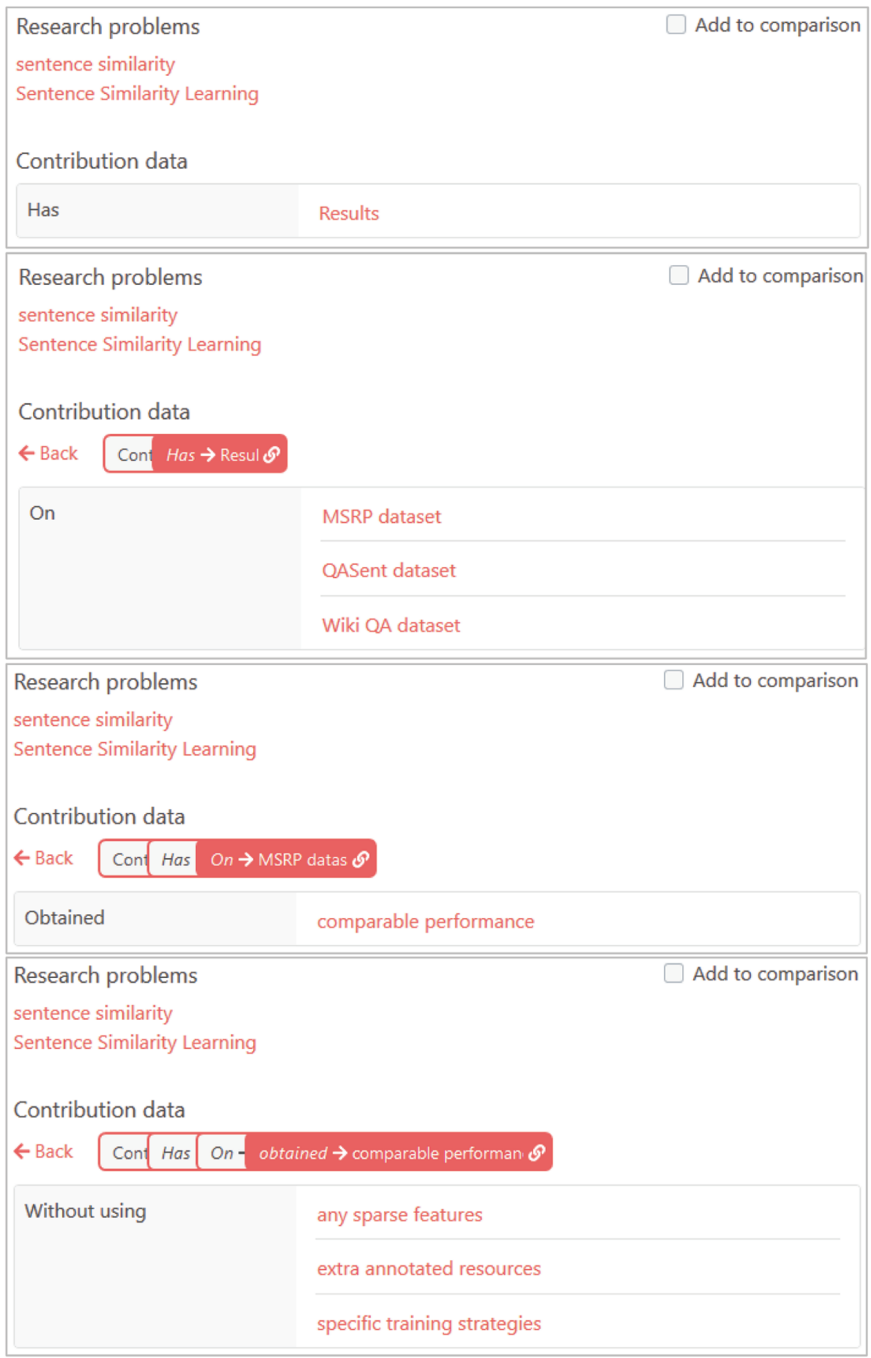

Figure 6. A RESULTS graph branch traversal in the ORKG until the last level.

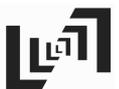

Figure 7. A NLPCONTRIBUTIONGRAPH Scheme Data Integration Use Case in the Open Research Knowledge Graph Digital Library. An automatically generated survey from a part of a knowledge graph of scholarly contributions over four articles using the NLPCONTRIBUTIONGRAPH scheme proposed in this work. This comparison was customized in the Open Research Knowledge Graph framework to focus only on the RESULTS information unit (the comparison is accessible online here https://www.orkg.org/orkg/c/kM2tUq).

As future directions, to realize a full-fledged KG in the context of the NCG scheme, there are a few IE modules that would need to be improved or added. This includes (1) improving the PDF parser (see Appendix 3 for challenges); (2) incorporating an entity and relation linking and normalization module; (3) merging phrases from the unstructured text with known ontologies (e.g. MEX (Esteves et al., 2015)) to align resources and thus ensure data interoperability and reusability; and (4) modeling inter-domain knowledge.

## Acknowledgments

This work was co-funded by the European Research Council for the project ScienceGRAPH (Grant agreement ID: 819536) and by the TIB Leibniz Information Centre for Science and Technology.

## Author contributions

Conceptualization: Jennifer D'Souza (jennifer.dsouza@tib.eu) and Sören Auer (auer@tib.eu); funding acquisition: Sören Auer; methodology: Jennifer D'Souza; original draft: Jennifer D'Souza; writing and reviewing final version: Jennifer D'Souza and Sören Auer.

## References

A reintroduction to our Knowledge Graph and knowledge panels. (2020). https://blog.google/products/search/about-knowledge-graph-and-knoswledge-panels/

Ammar, W., Peters, M.E., Bhagavatula, C., & Power, R. (2017). The AI2 system at SemEval-2017 Task 10 (ScienceIE): Semi-supervised end-to-end entity and relation extraction. SemEval@ACL.

Aryani, A., Poblet, M., Unsworth, K., Wang, J., Evans, B., Devaraju, A., Hausstein, B., Klas, C.-P., Zapilko, B., & Kaplun, S. (2018). A Research Graph dataset for connecting research data repositories using RD-Switchboard. Scientific Data, 5, 180099.

Auer, S. (2018). Towards an Open Research Knowledge Graph (Version 1) [Computer software]. Zenodo. https://doi.org/10.5281/zenodo.1157185

Augenstein, I., Das, M., Riedel, S., Vikraman, L., & McCallum, A. (2017). SemEval 2017 Task 10: ScienceIE—Extracting Keyphrases and Relations from Scientific Publications. SemEval@ACL.

Baas, J., Schotten, M., Plume, A., Côté, G., & Karimi, R. (2020). Scopus as a curated, high-quality bibliometric data source for academic research in quantitative science studies. Quantitative Science Studies, 1(1), 377–386.

Beltagy, I., Lo, K., & Cohan, A. (2019). SciBERT: A pretrained language model for scientific text. Proceedings of the 2019 Conference on Empirical Methods in Natural Language Processing and the 9th International Joint Conference on Natural Language Processing (EMNLP-IJCNLP), 3606–3611.

Birkle, C., Pendlebury, D.A., Schnell, J., & Adams, J. (2020). Web of Science as a data source for research on scientific and scholarly activity. Quantitative Science Studies, 1(1), 363–376.

Brack, A., D'Souza, J., Hoppe, A., Auer, S., & Ewerth, R. (2020). Domain-independent extraction of scientific concepts from research articles. European Conference on Information Retrieval, 251–266.

Burton, A., Koers, H., Manghi, P., La Bruzzo, S., Aryani, A., Diepenbroek, M., & Schindler, U. (2017). The data-literature interlinking service: Towards a common infrastructure for sharing data-article links. Program: electronic library and information systems, 51(1), 75–100. https://doi.org/10.1108/PROG-06-2016-0048

Buscaldi, D., Dessì, D., Motta, E., Osborne, F., & Reforgiato Recupero, D. (2019). Mining scholarly data for fine-grained knowledge graph construction. CEUR Workshop Proceedings, 2377, 21–30.

Camacho-Collados, J., & Pilehvar, M.T. (2017). On the role of text preprocessing in neural network architectures: An evaluation study on text categorization and sentiment analysis. ArXiv Preprint ArXiv:1707.01780.

Cho, K., Van Merriënboer, B., Gulcehre, C., Bahdanau, D., Bougares, F., Schwenk, H., & Bengio, Y. (2014). Learning phrase representations using RNN encoder-decoder for statistical machine translation. ArXiv:1406.1078.

Cimiano, P., Mädche, A., Staab, S., & Völker, J. (2009). Ontology learning. In Handbook on ontologies (pp. 245–267). Springer.

Constantin, A., Peroni, S., Pettifer, S., Shotton, D., & Vitali, F. (2016). The document components ontology (DoCO). Semantic Web, 7(2), 167–181.

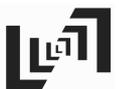

Devlin, J., Chang, M.-W., Lee, K., & Toutanova, K. (2018). Bert: Pre-training of deep bidirectional transformers for language understanding. ArXiv:1810.04805.

D'Souza, J., & Auer, S. (2020). NLPContributions: An Annotation Scheme for Machine Reading of Scholarly Contributions in Natural Language Processing Literature. In C. Zhang, P. Mayr, W. Lu, & Y. Zhang (Eds.), Proceedings of the 1st Workshop on Extraction and Evaluation of Knowledge Entities from Scientific Documents co-located with the ACM/IEEE Joint Conference on Digital Libraries in 2020, EEKE@JCDL 2020, Virtual Event, China, August 1st, 2020 (Vol. 2658, pp. 16–27). CEUR-WS.org. http://ceur-ws.org/Vol-2658/paper2.pdf

D'Souza, J., Hoppe, A., Brack, A., Jaradeh, M.Y., Auer, S., & Ewerth, R. (2020). The STEM-ECR Dataset: Grounding Scientific Entity References in STEM Scholarly Content to Authoritative Encyclopedic and Lexicographic Sources. LREC, 2192–2203.

Esteves, D., Moussallem, D., Neto, C.B., Soru, T., Usbeck, R., Ackermann, M., & Lehmann, J. (2015). MEX vocabulary: A lightweight interchange format for machine learning experiments. Proceedings of the 11th International Conference on Semantic Systems, 169–176.

Fisas, B., Ronzano, F., & Saggion, H. (2016). A Multi-Layered Annotated Corpus of Scientific Papers. LREC.

Fricke, S. (2018). Semantic scholar. Journal of the Medical Library Association: JMLA, 106(1), 145.

Ghaddar, A., & Langlais, P. (2018). Robust lexical features for improved neural network named-entity recognition. ArXiv:1806.03489.

GROBID. (2008). GitHub. https://github.com/kermitt2/grobid

Handschuh, S., & QasemiZadeh, B. (2014). The ACL RD-TEC: a dataset for benchmarking terminology extraction and classification in computational linguistics. COLING 2014: 4th International Workshop on Computational Terminology.

Hendricks, G., Tkaczyk, D., Lin, J., & Feeney, P. (2020). Crossref: The sustainable source of community-owned scholarly metadata. Quantitative Science Studies, 1(1), 414–427.

Huth, E.J. (1987). Structured abstracts for papers reporting clinical trials. American College of Physicians.

Jaradeh, M.Y., Oelen, A., Farfar, K.E., Prinz, M., D'Souza, J., Kismihók, G., Stocker, M., & Auer, S. (2019). Open Research Knowledge Graph: Next Generation Infrastructure for Semantic Scholarly Knowledge. KCAP, 243–246.

Jiang, M., D'Souza, J., Auer, S., & Downie, J.S. (2020). Targeting Precision: A Hybrid Scientific Relation Extraction Pipeline for Improved Scholarly Knowledge Organization. Proceedings of the Association for Information Science and Technology, 57(1).

Jinha, A.E. (2010). Article 50 million: An estimate of the number of scholarly articles in existence. Learned Publishing, 23(3), 258–263.

Johnson, R., Watkinson, A., & Mabe, M. (2018). The STM report. An Overview of Scientific and Scholarly Publishing. 5th Edition October.

Kononova, O., Huo, H., He, T., Rong, Z., Botari, T., Sun, W., Tshitoyan, V., & Ceder, G. (2019). Text-mined dataset of inorganic materials synthesis recipes. Scientific Data, 6(1), 1–11.

Kulkarni, C., Xu, W., Ritter, A., & Machiraju, R. (2018). An Annotated Corpus for Machine Reading of Instructions in Wet Lab Protocols. NAACL: HLT, Volume 2 (Short Papers), 97–106. https://doi.org/10.18653/v1/N18-2016

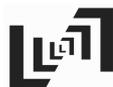

Kuniyoshi, F., Makino, K., Ozawa, J., & Miwa, M. (2020). Annotating and Extracting Synthesis Process of All-Solid-State Batteries from Scientific Literature. LREC, 1941–1950.

Lample, G., Ballesteros, M., Subramanian, S., Kawakami, K., & Dyer, C. (2016). Neural architectures for named entity recognition. ArXiv Preprint ArXiv:1603.01360.

Landhuis, E. (2016). Scientific literature: Information overload. Nature, 535(7612), 457–458.

Liakata, M., Saha, S., Dobnik, S., Batchelor, C., & Rebholz-Schuhmann, D. (2012). Automatic recognition of conceptualization zones in scientific articles and two life science applications. Bioinformatics, 28(7), 991–1000.

Liakata, M., Teufel, S., Siddharthan, A., & Batchelor, C.R. (2010). Corpora for the Conceptualisation and Zoning of Scientific Papers. LREC.

Lin, D.K., & Pantel, P. (2002). Concept discovery from text. COLING 2002: The 19th International Conference on Computational Linguistics.

Luan, Y., He, L., Ostendorf, M., & Hajishirzi, H. (2018). Multi-Task Identification of Entities, Relations, and Coreference for Scientific Knowledge Graph Construction. EMNLP.

Luan, Y., Ostendorf, M., & Hajishirzi, H. (2017). Scientific information extraction with semi-supervised neural tagging. ArXiv:1708.06075.

Mysore, S., Jensen, Z., Kim, E., Huang, K., Chang, H.-S., Strubell, E., Flanigan, J., McCallum, A., & Olivetti, E. (2019). The Materials Science Procedural Text Corpus: Annotating Materials Synthesis Procedures with Shallow Semantic Structures. Proceedings of the 13th Linguistic Annotation Workshop, 56–64.

Noy, N., Gao, Y., Jain, A., Narayanan, A., Patterson, A., & Taylor, J. (2019). Industry-scale knowledge graphs: Lessons and challenges. Queue, 17(2), 48–75.

Oelen, A., Jaradeh, M.Y., Farfar, K.E., Stocker, M., & Auer, S. (2019). Comparing research contributions in a scholarly knowledge graph. CEUR Workshop Proceedings 2526 (2019), 2526, 21–26.

Oelen, A., Jaradeh, M.Y., Stocker, M., & Auer, S. (2020). Generate FAIR Literature Surveys with Scholarly Knowledge Graphs. Proceedings of the ACM/IEEE Joint Conference on Digital Libraries in 2020, 97–106. https://doi.org/10.1145/3383583.3398520

Pertsas, V., & Constantopoulos, P. (2017). Scholarly Ontology: Modelling scholarly practices. International Journal on Digital Libraries, 18(3), 173–190.

Qi, P., Zhang, Y.H., Zhang, Y.H., Bolton, J., & Manning, C.D. (2020). Stanza: A Python Natural Language Processing Toolkit for Many Human Languages. Proceedings of the 58th Annual Meeting of the Association for Computational Linguistics: System Demonstrations. https://nlp.stanford.edu/pubs/qi2020stanza.pdf

Soldatova, L.N., & King, R.D. (2006). An ontology of scientific experiments. Journal of the Royal Society, Interface, 3 11, 795–803.

Sollaci, L.B., & Pereira, M.G. (2004). The introduction, methods, results, and discussion (IMRAD) structure: A fifty-year survey. Journal of the Medical Library Association, 92(3), 364.

Teufel, S., Carletta, J., & Moens, M. (1999). An annotation scheme for discourse-level argumentation in research articles. Proceedings of the Ninth Conference on European Chapter of ACL, 110–117.

Teufel, S., Siddharthan, A., & Batchelor, C. (2009). Towards discipline-independent argumentative zoning: Evidence from chemistry and computational linguistics. EMNLP: Volume 3, 1493–1502.

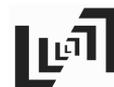

Vogt, L., D'Souza, J., Stocker, M., & Auer, S. (2020). Toward representing research contributions in scholarly knowledge graphs using knowledge graph cells. Proceedings of the ACM/IEEE Joint Conference on Digital Libraries in 2020, 107–116.

Vrandečić, D., & Krötzsch, M. (2014). Wikidata: A free collaborative knowledgebase. Communications of the ACM, 57(10), 78–85.

Wang, B.L., Lu, W., Wang, Y., & Jin, H.X. (2018). A neural transition-based model for nested mention recognition. ArXiv:1810.01808.

Wang, K.S., Shen, Z.H., Huang, C.Y., Wu, C.-H., Dong, Y.X., & Kanakia, A. (2020). Microsoft academic graph: When experts are not enough. Quantitative Science Studies, 1(1), 396–413.

Wilkinson, M.D., Dumontier, M., Aalbersberg, Ij. J., Appleton, G., Axton, M., Baak, A., Blomberg, N., Boiten, J.-W., da Silva Santos, L.B., Bourne, P.E., & others. (2016). The FAIR Guiding Principles for scientific data management and stewardship. Scientific Data, 3(1), 1–9.

Zhou, J., Cao, Y., Wang, X.G., Li, P., & Xu, W. (2016). Deep recurrent models with fast-forward connections for neural machine translation. Transactions of the Association for Computational Linguistics, 4, 371–383.

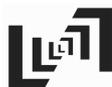

# Appendix

## 1. NLPCONTRIBUTIONGRAPH: Parent Node Names

While the node names for the top two levels of the NLPCONTRIBUTIONGRAPH graph data annotations are predefined, the remaining node names are unspecified and are left as data-driven decisions. Higher up in the graph, in particular, they are often selected as the names from titles of sections and subsections in the article. Sometimes, however, they are also selected from the running text. In this case the phrase has to serve the name of a conceptual reference for an idea described more in detail. In Figure 8 we illustrate this with an example.

```
"employ" : {
  "Stack - LSTM" : {
    "to represent" : {...},
    "has" : {...},
    "incorporate" : {
      "characterlevel LSTM" : {
        "to capture" : "morphological information",
        "help deal with" : "out - of - vocabulary
            problem of neural models"
      },
      "from sentence" : "Based on the observation that
          letter - level patterns such as capitalization
          and prefix can be beneficial in detecting
          mentions , we incorporate a characterlevel
          LSTM to capture such morphological information
          . Meanwhile , this character - level component
          can also help deal with the out - of -
          vocabulary problem of neural models ."
    }
  }
}
```

Figure 8.   Illustration of a parent node name called 'character-level LSTM' serving a conceptual reference selected from the article's running text as opposed to the section names. The figure is part of the contribution from the article (B. Wang et al., 2018). Essentially, for such encapsulation when it exists, coreference is applied for the child-node nesting (consider the coreference between 'we incorporate a character-level LSTM to capture' in sentence 1 and 'this character-level component can also help' in sentence 2).

## 2. Improved Phrasal Granularity during Adjudication

Between the pilot and adjudication stages, the data quality was far improved in various aspects. One of these aspects included modeling a more consistent phrasal granularity, as the second data element in the NLPCONTRIBUTIONGRAPH data model. This is comparatively illustrated between data from the pilot and adjudication stages in Figure 9.

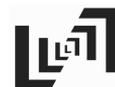

```
{
  "has" : {
    "Results" : {
      "on" : {
        "CoNLL" : {
          "significantly outperforms" : ["models
              that use extensive sets of hand
              -crafted features", "system of (Luo et
              al., 2015) that uses NE and Entity
              Linking annotations to jointly
              optimize the performance on both
              tasks", {"from sentence" : "First, we
              observe that our model significantly
              outperforms models that use extensive
              sets of handcrafted features (Ratinov
              and Roth, 2009; Lin and Wu, 2009) as
              well as the system of (Luo et al.,
              2015) that uses NE and Entity Linking
              annotations to jointly optimize the
              performance on both tasks."}],
```

(a) Pilot stage annotations

```
{
  "has" : {
    "Results" : {
      "on" : {
        "CoNLL" : {
          "significantly outperforms" : {
            "models" : {
              "use" : "extensive sets of
                  handcrafted features"
            },
            "from sentence" : "First , we
                observe that our model
                significantly outperforms models
                that use extensive sets of
                handcrafted features ) as well as
                the system of Standard deviation
                on the test set is reported in
                2015 ) that uses NE and Entity
                Linking annotations to jointly
                optimize the performance on both
                tasks . "
```

(b) Adjudication stage annotations

Figure 9.  Figures (a) and (b) depicts the modeling of part of a Results information unit from a scholarly article (Ghaddar & Langlais, 2018) in the pilot and the adjudication stages, respectively.

### 3. Scanned PDF Extraction ... garbage in garbage out

The Grobid (GROBID, 2008) parser, while a state-of-the-art tool, produces noise in its parsed plain-text. For instance, while it retains no citations, tables, or figures in the parsed output, it does retain captions. This becomes a source of noise as often caption text is intermingled within the section text. Further, if the parser could not decode symbols, including math symbols such as α, β, or ε or formulae, they are simply a "?" in the parsed plain text. Also in many cases, entire paragraphs of the original paper content are lost, in which case such information, even if relevant, cannot be modeled by NlpContributionGraph. E.g. in "Deep Recurrent Models with Fast-Forward Connections for Neural Machine Translation (Zhou et al., 2016)," a full page of Results information content is not present in the plaintext parsed output and hence our annotations do not include the results written in the missing task. Finally, the parsed text can also be produced out of order. E.g. in reference paper (Camacho-Collados & Pilehvar, 2017), the section header "Experiment 1: Preprocessing effect" is seen sentence 94 in the plaintext, however, its following sentence in the original text is seen as sentence 79.

Overall, the parsing function tends to be noisy around tables, figures, and footnotes. In which case entire sections of content is not returned, or is noisy with intermingled captions, or out of sequence.

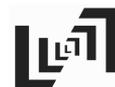